\pdfoutput=1

\documentclass[11pt]{article}

\usepackage{ACL2023}

\usepackage{times}
\usepackage{latexsym}

\usepackage{tabularx}
\usepackage{arydshln}
\usepackage[cjk]{kotex}

\usepackage[T1]{fontenc}

\usepackage[utf8]{inputenc}

\usepackage{microtype}
\usepackage{graphicx}

\usepackage{inconsolata}

\usepackage{todonotes}

\title{K-UniMorph: Korean Universal Morphology and its Feature Schema}

\author{
Eunkyul Leah Jo$^{1*}$~~~ 
Kyuwon Kim$^{1,2*}$~~~ 
Xihan Wu$^{1}$\thanks{~~Equally contributed authors.}~~~ 
{KyungTae Lim}$^{3}$\\
\textbf{Jungyeul Park}$^{1}$~~~ 
\textbf{Chulwoo Park}$^{4}$\\
$^{1}$The University of British Columbia, Canada~~ $^{2}$Seoul National University, South Korea \\
$^{3}$SeoulTech \& Teddysum, South Korea~~ $^{4}$Anyang University, South Korea \\
{\tt\{eunkyul,wuxihan\}@student.ubc.ca}~~~ {\tt guwon0406@snu.ac.kr}~~~ {\tt ktlim@seoultech.ac.kr} \\
{\tt jungyeul@mail.ubc.ca}~~~ {\tt cwpa@anyang.ac.kr}\\
}

\begin{document}

\maketitle

\begin{abstract}
We present in this work a new Universal Morphology dataset for Korean. Previously, the Korean language has been underrepresented in the field of morphological paradigms amongst hundreds of diverse world languages. Hence, we propose this Universal Morphological paradigms for the Korean language that preserve its distinct characteristics. For our K-UniMorph dataset, we outline each grammatical criterion in detail for the verbal endings, clarify how to extract inflected forms, and demonstrate how we generate the morphological schemata. This dataset adopts morphological feature schema from \citet{sylakglassman-EtAl:2015} and \citet{sylakglassman:2016} for the Korean language as we extract inflected verb forms from the Sejong morphologically analyzed corpus that is one of the largest annotated corpora for Korean.
During the data creation, our methodology also includes investigating the correctness of the conversion from the Sejong corpus. Furthermore, we carry out the inflection task using {three different}
Korean word forms: letters, syllables and morphemes. Finally, we discuss and describe future perspectives on Korean morphological paradigms and the dataset.
\end{abstract}

\section{Introduction}

The Universal Morphology (UniMorph) project is a collaborative effort providing broad-coverage morphological paradigms for diverse world languages \citep{mccarthy-etal-2020-unimorph,kirov-EtAl:2018:LREC}.
UniMorph consists of a lemma and bundle of morphological features related to a particular inflected word form as follows, for example:

\begin{center}
{나서다}\textit{naseoda} {나섰다}\textit{naseossda} \textsc{v;decl;pst}
\end{center}
\noindent  where 나서다\textit{naseoda} is the lemma form and 나섰다\textit{naseossda} 
(`became') is the inflected form with \textsc{v;decl;pst} (verb, declarative, and past tense) as morphological schema.

It started in 2016 as a SIGMORPHON shared task \citep{cotterell-EtAl:2016:SIGMORPHON} for the problem of morphological reinflection, and it introduced morphological datasets for 10 languages.  
The inflection task, using the given lemma with its part-of-speech to generate a target inflected form, has been continued through the years: CoNLL–SIGMORPHON 2017 Shared Task \citep{cotterell-EtAl:2017:CoNLL}, CoNLL–SIGMORPHON 2018 Shared Task \citep{cotterell-EtAl:2018:CoNLL}, SIGMORPHON 2019 Shared Task \citep{mccarthy-etal-2019-sigmorphon}, SIGMORPHON 2020 Shared Task \citep{gorman-etal-2020-sigmorphon} and SIGMORPHON 2021 Shared Task \citep{pimentel-ryskina-etal-2021-sigmorphon}. 
However, the Korean language has not been a part of the shared task because of the lack of the dataset. 

Nonetheless, although rarely, morphological paradigms for Korean have been explored in the context of computational linguistics.
\citet{yongkyoon:1993} defined the inflectional classes for verbs in Korean using word-and-paradigm (WP) \citep{hockett:1954} approaches. 
His fifteen classes of the verb which can be joined with seven different types of verbal endings, are based on inflected forms of the verb. 
\citet{seokjoon:1999} systematized the list of final endings and their properties, which are also used as conjunctive endings in Korean.
Otherwise, properties of verbs such as mood, tense, voice, evidentiality, interrogativity have been extensively studied in Korean linguistics independently: for example, \textit{inter alia}, 
tense \citep{byungsun:2003}, 
grammatical voice \citep{chulwoo:2007}, 
interaction of tense–aspect–mood marking with modality \citep{jaemog:1998}, 
evidentiality \citep{donghoon:2008}, and
interrogativity \citep{donghoon:2011}. 

In continuation of the efforts, this paper proposes a new Universal Morphology dataset for Korean. 
We adopt morphological feature schema from \citet{sylakglassman-EtAl:2015} and \citet{sylakglassman:2016} for the Korean language and extract inflected verb forms from the Sejong morphologically analyzed corpus over 0.6M sentences with 9.5M words. 
We set the criteria in detail by explaining how to extract inflected verbal forms (Section~\ref{feature-schema}), and carry out the inflection task using different Korean word forms such as letter, syllable and morpheme (Section~\ref{experiments}).
Finally, we discuss future perspectives on a Korean UniMorph dataset (Section~\ref{discussion}).

\section{UniMorph Features Schema} \label{feature-schema}

Verbal endings in the inflected forms of the predicate has been considered {as} still {being in} the part of the word {as proposed} in {several grammar} formalisms for Korean such as lexicalized tree adjoining grammars \citep{park:2006}, head driven phrase structure grammars \citep{ko:2010}, and combinatory categorial grammars \citep{kang:2011} {in contrast to government and binding (GB) theory \citep{chomsky:1981,chomsky:1982} for Korean in which the entire sentence depends on separated verbal endings}.
This idea goes back to Maurice Gross's lexicon grammars \citep{gross:1975}, and his students who worked on a descriptive analysis of Korean in which the number of predicates in Korean could be fixed by generating possible inflection forms: \textit{e.g.} \citet{pak:1987,nho:1992,nam:1994,shin:1994,park:1996,chung:1998,han:2000}.
{However,} we have separated the postposition from the substantive such as noun phrases instead of keeping themselves together. 
{Therefore,} with the {current} Korean dataset, we {decide to} annotate morphological data for verbs (\textsc{v}). 

Table~\ref{morphological-schema-verb} shows the morphological schema for Korean UniMorph where we adopt features from \citet{sylakglassman-EtAl:2015} and \citet{sylakglassman:2016} for the Korean language.
{In addition to the features schema,} we consider following these four different types of verbal endings{, in which they convey} grammatical meanings for the predicate: sentence final ending (\texttt{ef}),  non-final ending (\texttt{ep}),  conjunctive ending (\texttt{ec}), and modifier ending (\texttt{etm}).

\begin{table*}[ht]
\centering
\begin{tabularx}{.98\textwidth}{l l X} \hline

Evidentiality 
& \textsc{hrsy} & hearsay: 일\textit{il} (`work')/NNB 이\textit{i} (`\textsc{cop}')/VCP + \textbf{래\textit{lae} (`\textsc{hrsy}')/EF} (`happen')\\
& \textsc{infer} & inferred: 괜찮\textit{gwaenchanh} (`fine')/VA + \textbf{겠\textit{gess} (`\textsc{infer}')/EP} + 다\textit{da} (`\textsc{decl}')/EF  \\
\hline 

Interrogativity & 
\textsc{decl} &  declarative: 모이\textit{moi} (`gather')/VV + \textbf{ㄴ다\textit{nda} (`\textsc{decl}')/EF}  \\
& \textsc{int} & interrogative: 배우\textit{baeu} (`study')/VV + \textbf{는가\textit{neunga} (`\textsc{int}')/EF} \\
\hline

Mood 
& \textsc{real} & realis:  얻\textit{eod} (`get')/VV + \textbf{은\textit{eun} (`\textsc{real}')/ETM} \\

& \textsc{irr} & irrealis: 잊\textit{ij} (`forget')/VV + \textbf{을\textit{eul} (`\textsc{irr}')/ETM}   \\

& \textsc{purp} & general purposive: 달래\textit{dallae} (`appease')/VV + \textbf{려고\textit{lyeogo} (`\textsc{purp}')/EC} \\

& \textsc{oblig} & obligative: 이어지\textit{ieoji} (`connect')/VV + \textbf{어야\textit{eoya} (`\textsc{oblig}')/EC} (`should be connected')\\
\hline



Tense & 
\textsc{prs} &  present: 들리\textit{deulli} + (`hear')/VV + \textbf{ㄴ다\textit{nda} (`\textsc{prs},\textsc{decl}')/EF}  \\
& \textsc{pst} & past: 나타나\textit{natana} (`appear')/VV + \textbf{았\textit{ass} (`\textsc{pst}')/EP} + 다\textit{da} (`\textsc{decl}')/EF   \\
\hline

Voice & 
\textsc{caus} & causative: 보이\textit{boi} (`show')/VV + \textbf{게\textit{ge} (`\textsc{caus}')/EC} \\ 
& \textsc{pass} & passive: \textbf{잡히\textit{jabhi} (`be caught')/VV} + 었\textit{eoss} (`\textsc{pat}')/EP + 다\textit{da} (`\textsc{decl}')/EF   \\
\hline

\end{tabularx}
\caption{Korean UniMorph schema for verbs: {\texttt{vv} for verb, \texttt{va} for adjective, \texttt{vcp} for copula, and \texttt{nnb} for bound noun}, }
\label{morphological-schema-verb}
\end{table*}

\paragraph{Evidentiality} It is a grammatical category that reflects the source of information that a speaker conveys in a proposition. 
It is often expressed through morphological markers such as sentence final endings (\texttt{ef}) 대\textit{dae}, 내\textit{nae}, and 래\textit{lae} bring in  hearsay (\textsc{hrsy}), and non-final endings (\texttt{ep}) 겠\textit{gess} introduce inferred (\textsc{infer}).
Since the suffix for the quotative (\textsc{quot}) is denoted with a postposition (\texttt{jkq}) in Korean instead of the verbal ending, it is excluded from the current set of schemata.

\paragraph{Interrogativity} It indicates either to express a statement (\textsc{decl}) or a question (\textsc{int}). 
We consider all sentence final ending (\texttt{ef}) ended with 다\textit{da} as declarative \textsc{decl}, and sentence final ending (\texttt{ef}) included 가\textit{ga} and 까\textit{kka} as interrogative \textsc{int}.

\paragraph{Mood} The grammatical mood of a verb indicates modality on a verb by  the morphological marking. Realis (\textsc{real}) and irrealis (\textsc{irr}) are represented by a verbal modifier ending  (also known as an adnominal ending) (\texttt{etm}), ㄴ\textit{n} and ㄹ\textit{l}, respectively.
The usage of adnominal endings consists of (i) collocation such as 인한\textit{inhan}, 치면\textit{chimyeon}, 대한\textit{daehan}, (ii) modifiers and (iii) relative clauses. 
Realis and irrealis are concerned with regardless of modifiers or relative clauses. 
General purposive (\textsc{purp}) is decided by 려고\textit{lyeogo} and  하러\textit{haleo}, and obligative (\textsc{oblig}) is introduced by 야\textit{ya}. 
It is worthwhile to note that we do not consider indicative (\textsc{ind}) because we specify declarative \textsc{decl}.

\paragraph{Tense} It refers to the time frame in which a verb's action or state of being occurs. 
Non-final endings (\texttt{ep}) such as 았\textit{ass} and 었\textit{eoss} and
final endings (\texttt{ef}) such as ㄴ다\textit{nda} 는다\textit{neunda} can represent the past (\textsc{past}) and  the present (\textsc{prs})  tenses, repectively.
Since the future tense (\textsc{fut}) has been considered as irrealis (\textsc{irr}) in Korean, we don't annotate it here.

\paragraph{Voice} 
We deduce the passive (\textsc{pass}) from the verb stem instead of the verbal ending such as \textit{jab-hi} (`be caught'). Whereas the verb \textit{jab} (`catch') and the passive suffix \textit{hi} might be segmented, the current criteria of the Sejong corpus combines them together as a single morpheme. 
이히리기\textit{i}, \textit{hi}, \textit{li}, \textit{gi} are verbal endings known for both the passive and the causative. If the verb has a verbal ending 게\textit{ge} such as \texttt{verb stem}+\{이\textit{i}|히\textit{hi}|리\textit{li}|기\textit{gi}\}+게\textit{ge} \{하\textit{ha}|만들\textit{mandeul} (`make')\}, then it is causative (\textsc{caus}), otherwise passive (\textsc{pass}). 

\paragraph{Other schema}
For politeness, we introduce only polite (\textsc{pol}) using the non-final ending (\texttt{ep}) 시\textit{si} as the direct encoding of the speaker-addressee relationship \citep[p.276]{brown-levinson:1987}. 
Lastly, since we are not able to deduce the valency of the verb from morphemes, we do not include \textsc{intr} (intransitive), \textsc{tr} (transitive) and \textsc{ditr} (ditransitive). 
However, we leave {them} for future work because the valency might still be valid morphological feature schemata for Korean.

\section{Experimental Results} \label{experiments}

\subsection{Data creation}
We prepare the data by extracting inflected verb forms from the Sejong morphologically analyzed corpus (\texttt{sjmorph}) over 676,951 sentences with  7,835,239 eojeols (word units separated by space) which represent 9,537,029 tokens. 
We are using the same training/dev/test data split that \citet{park-tyers:2019:LAW} proposed for Korean part of speech (POS) tagging. 
However, the current \texttt{sjmorph} doesn't contain POS labels for the eojeol (the word). Instead, it contains the sequence of POS labels for morphemes as follows: 

\begin{center}
\resizebox{.48\textwidth}{!}{
{나섰다}\textit{naseossda}~~~ {나서}\textit{naseo}/VV+{었}\textit{eoss}/EP+{다}\textit{da}/EF
}
\end{center}

\noindent where it {contains} only each morpheme's POS label: a {verb} {나서}\textit{naseo} (`become'), a non-final ending {었}\textit{eoss} (`\textsc{pst}'), and a final ending {다}\textit{da} (`\textsc{decl}'), and it does not show whether the word {나섰다}\textit{naseossda} (`became') is a verb.
Previous works \citep{petrov-das-mcdonald:2012:LREC,park-hong-cha:2016:PACLIC,park-tyers:2019:LAW,kim-colineau:2020:LREC} propose a {partial} mapping table between Sejong POS (and the sequence of Sejong POSs) (XPOS) and Universal POS (UPOS) labels where UPOS represents {the grammatical category} of the word. 
However, no study has presented the correctness of their conversion rules. 
Therefore, we utilize \texttt{UD\_Korean-GSD} \citep{mcdonald-EtAl:2013:ACL} in Universal Dependencies \citep{nivre-EtAl:2016:LREC,nivre-EtAl:2020:LREC} that provides Sejong POS(s) and Universal POS labels for each word. {Nevertheless, we observed several critical POS annotation errors in \texttt{UD\_Korean-GSD}. For this reason, we proceeded to revise \texttt{GSD}'s  Sejong POS(s) and Universal POS to evaluate our criteria of getting verbs (inflected forms and their lemmas) from \texttt{sjmorph}. }
This approach involved randomly selecting 300 sentences from the \texttt{GSD} and manually revising their POS labels based on the Sejong POSs. For thorough verification, they were examined by our linguist for over 60 hours over 3 weeks. 
The main places of error that we noticed were how words for {proper nouns}
were labeled as \texttt{NOUN} even with its XPOS of {proper nouns} (\texttt{NNP}). They were corrected to the UPOS label of \texttt{PROPN}. Another common place of error was how the dataset recognized and labeled words according to their roles as constituent parts of the sentence they are in{,} instead of the word's own category. 
For example, the temporal nouns was usually annotated as \texttt{ADV} instead of \texttt{NOUN}.
{W}e changed this mislabeling by acknowledging the word itself, separate from the sentence. Again, the Sejong POS labels were revised based on the criteria of the Sejong corpus.
After correcting {738} words for Sejong POS labels and {705} words for Universal POS labels {from 300 sentences in} the development file, we trained the sequence of Sejong POS labels {using semi-supervised learning} to predict the Universal POS label for each word. 
Among 3674 predictions, there were only 332 UPOS prediction errors, and {an error scarcely occurs for \texttt{VERB}} labels, which we {attempted} to extract from \texttt{sjmorph}. Therefore, we consider this current error rate for the verb to be {negligible.}
Finally, we extract {244,871} inflected verbal forms for {43,959} lemma types from \texttt{sjmorph}. 
{Then,} {we remove all duplicated items from train+dev datasets compared to the test dataset. {In Table~\ref{stat} is the} brief statistics of the current dataset.}

\begin{table}
\centering
\begin{tabular}{ r | ccc} \hline
     &  train & dev & test \\ \hline
lemma     & 41,631 &7505 & 7595 \\
inflected     & 197,774 & 19,251 & 27,846 \\\hline
\end{tabular}
\caption{Statistics of Korean UniMorph} \label{stat}
\end{table}



\subsection{Morphological reinflection} \label{morphological-reinflection}
The goal of the morphological reinflection task creates the generative function of morphological schema to produce the inflected form of the given word. 
{For Korean,  {we use} {나서다}\textit{naseoda} and \textsc{v;decl;pst} {to predict} {나섰다}\textit{naseossda} by using the composition of alphabet letters (\textsc{l}), syllables (\textsc{s}) and morphemes (\textsc{m}) of the word as shown in Table~\ref{inflection-table}. 
{The word is decomposed into the sequence of consonants and vowels by \texttt{Letter}, the sequence of units constructed with two or three letters by \texttt{syllable}, and the sequence of morphological units by \texttt{morpheme}.
The conversion from the target form of each representation to the surface form and vice versa are straightforward in technical terms.}
}

\begin{table}
    \centering
\resizebox{.45\textwidth}{!}
{
\begin{tabular}{r| cc}\hline
        & Source & Target \\\hline
\texttt{letter} (\textsc{l})  & ㄴㅏㅅㅓㄷㅏ& ㄴㅏㅅㅓㅆㄷㅏ \\
\texttt{syllable}  (\textsc{s})& 나서다     &  나섰다\\
\texttt{morpheme}  (\textsc{m}) &  나서다 & 나서었다 \\ \hdashline
surface form &나서다\textit{naseoda} & 나섰다\textit{naseossda} \\\hline
\end{tabular}
}
\caption{Example of the surface form and its different representation using letters, syllables and morphemes.
}\label{inflection-table}
\end{table}

For our task, we use the baseline system from The CoNLL–SIGMORPHON 2018 Shared Task \citep{cotterell-EtAl:2018:CoNLL}.\footnote{\url{https://github.com/sigmorphon/conll2018}}
The system uses alignment, span merging and rule extraction to predict the set of all inflected forms of a lexical item \citep{durrett-denero:2013:NAACL-HLT}. 
We also build a basic neural model using \texttt{fairseq}\footnote{\url{https://github.com/facebookresearch/fairseq}} \citep{ott-etal-2019-fairseq} and Transformer \citep{vaswani-EtAl:2017:NIPS}. 
Table~\ref{experiment-results} shows the experimental results for Korean UniMorph using the three different representation forms.
{It is notable that the morpheme forms outperform the other surface representation forms such as by letters and syllables of the word. This is because morpheme forms imply lemma forms for both source and target data.}
While the average number of inflected forms per lemma is 8.285, there are 22 verb lemmas that have more than 400 different inflected forms. 
The average number of inflected forms per lemma and morphological feature pair is also 5.634, {and this makes Korean} difficult to predict the inflected form.

\begin{table}
    \centering
\begin{tabular}{r | ccc}\hline
& \textsc{l} & \textsc{s} & \textsc{m} \\ \hline
baseline     & 26.88 & 27.75 & 31.29 \\
neural     & 51.97 & 49.72 & 54.26 \\\hline
\end{tabular}
\caption{{Experimental results (accuracy)}}
    \label{experiment-results}
\end{table}

\begin{table}
    \centering
\resizebox{.47\textwidth}{!}{
    \begin{tabular}{c | cc} \hline
& UniMorph 4.0 Korean & K-UniMorph \\ \hline
Evide. & - & 
        \textbf{\textsc{hrs}}, \textbf{\textsc{infer}} \\
Finit. & \textsc{fin}, \textsc{nfin} & - \\
Inter. & \textsc{decl}, \textsc{int}, \textsc{imp} & 
        \textsc{decl}, \textsc{int}\\
Mood &  \textsc{cond}, \textsc{purp} & 
        \textbf{\textsc{real}}, \textbf{\textsc{irr}}, \textsc{purp}, \textbf{\textsc{oblig}}\\
Tense & \textsc{prs}, \textsc{pst}, \textsc{fut} & 
        \textsc{prs}, \textsc{pst}\\
Voice & \textsc{caus} & 
        \textsc{caus}, \textbf{\textsc{pass}}\\  
Polit. & \textsc{form}, \textsc{inform}, \textsc{pol} \textsc{elev} & 
        \textsc{pol} \\
Per. & \textsc{1}, \textsc{2} & - \\
Num. & \textsc{pl} & - \\
\hline
    \end{tabular}
}
\caption{Feature schema comparison between UniMorph 4.0 Korean K-UniMorph.
}
    \label{table-comparison}
\end{table}

\begin{table*}[ht]
\begin{tabularx}{\textwidth}{l l X} \hline

Core case
& \textsc{nom} & nominative which marks the subject of a verb: 병원\textit{byeongwon} (`hospital')/NNG +  \textbf{이\textit{i} (`\textsc{nom}')/JKS} \\
& \textsc{acc} & accusative  which marks the object of a verb: 원인\textit{wonin} (`cause')/NNG + \textbf{을\textit{eul} (`\textsc{acc}')/JKO}  \\ \hdashline

Non-core, non-local case
& \textsc{dat} & dative which marks the indirect object: 국민\textit{gugmin} (`people')/NNG + 에게\textbf{\textit{ege} (`\textsc{dat}')/JKB}
\\ 
 & \textsc{gen} & genitive which marks the possessor: 사회\textit{sahoe} (`society')/NNG + \textbf{의\textit{ui} (`\textsc{gen}')/JKG} \\

& \textsc{ins} & instrumental which marks means by which an action occurred: 대리석\textit{daeliseog} (`marble')/NNG + \textbf{으로\textit{eulo} (`\textsc{ins}')/JKB}  \\ 

& \textsc{com} &  comitative which marks the accompaniment: 망치\textit{mangchi} (`hammer')/NNG + \textbf{와\textit{wa} (`\textsc{com}')/JC}\\
& \textsc{voc} & vocative which indicate the direct form of address: 달\textit{dal} (`moon')/NNG + \textbf{아\textit{a} (`\textsc{voc}')/JKV} \\ \hdashline

Local case 
& \textsc{all} & allative which marks a type of locative grammatical case: 길\textit{gil} (`road')/NNG + \textbf{로\textit{lo} (`\textsc{all}')/JKB} \\ 

& \textsc{abl} & ablative which expresses motion away from something: 밑\textit{mit} (`bottom')/NNG + \textbf{에서부터\textit{eseobuteo} (`\textsc{abl}')/JKB} \\ 

\hline

Comparison 
& \textsc{cmpr} &  comparative: 예상\textit{yesang} (`expectation')/NNG + \textbf{보다\textit{boda} (`\textsc{cmpr}')/JKB} \\
\hline

Information structure  & \textsc{top}& topic which is what is being talked about: 사람\textit{salam} (`people')/NNG + \textbf{은\textit{eun} (`\textsc{top}')/JX} \\
\hline
\end{tabularx} 
\caption{{Korean UniMorph schema for nouns.}}
\label{morphological-schema-noun}
\end{table*}

\subsection{Comparison with UniMorph 4.0 Korean}
UniMorph 4.0 \citep{batsuren-etal-2022-unimorph} includes a Korean dataset, which provides 2686 lemma and 241,323 inflected forms that are automatically extracted from Wiktionary. It is mainly comprised of adjectives and verbs with totals of 52,387 and 188,821, respectively.\footnote{The counts are short of some numbers because the errors, 92 forms without morphological schema, are excluded.}
Thoroughly, we inspected the verbs in UniMorph 4.0 Korean to compare with K-UniMorph:
Among the 152,454 inflected forms of verbs in UniMorph 4.0 Korean, there are only 16,489 forms that appear in 9.5M words of the Sejong corpus, and 135,965 forms (89.18\%) that never occur.
UniMorph 4.0 Korean annotated all verbs (\textsc{v}) as \textsc{fin} and all participles (\textsc{v.cptp}) as \textsc{nfin}. 
We can consider adding \textsc{fin} for all verbs endings with \texttt{ef} (final verbal endings) and \textsc{nfin} for all verbs ending with \texttt{etm} (adnominal endings, which are utilized for relative clauses, modifiers, and a part of collocations).
To inspect this, UniMorph 4.0 Korean provides the imperative-jussive modality \textsc{imp} which consists of \textsc{1}\texttt{;}\textsc{pl} and \textsc{2}, but it seems that Number (\textsc{pl}) occurs only with \textsc{1} (Person). 
While K-UniMorph considers only 시\textit{si} (an honorific for the agent) as \textsc{pol},  UniMorph 4.0 Korean uses \textsc{elev} for 시\textit{si}, and \textsc{pol} comes from verbal endings 요\textit{yo} and 습니다\textit{seubnida} with either \textsc{form} or \textsc{infm}. 
However, \textsc{form.elev} is to elevate the referent. Therefore, it should be with \textsc{imp;2$|$3} and instead, \textsc{form.humb} can be introduced with \textsc{imp;1} for 습니다\textit{seubnida}, and \textsc{infm.elev$|$infn.humb} for 요\textit{yo}. Hence, K-UniMorph provides a richer feature schema based on linguistics analysis. Table~\ref{table-comparison} summarises the different usage of the feature schema between UniMorph 4.0 Korean K-UniMorph.

\section{Discussion and Future Perspectives} \label{discussion}

We have dealt with UniMorph schema for verbs, {and obtained experimental results for the morphological reinflection task using the different representation forms of the word.
Nouns in Korean have been considered by separating postposition from the lemma of the noun instead of keeping themselves together (\textit{e.g.} 프랑스\textit{peulangseu} (`France') and 의\textit{ui} (`\textsc{gen}') instead of 프랑스의\textit{peulangseuui}) in several grammar formalisms for Korean. 
However, in addition to exogenously given interests such as \textit{inflection in context},\footnote{\url{https://sigmorphon.github.io/sharedtasks/2018/task2/}} recent studies insist the functional morphemes including both verbal endings and postpositions in Korean should be treated as part of a word, with the result that their categories do not require to be assigned individually in a syntactic level \citep{park-kim:2023}. 
Accordingly, it would be more efficient to assign the syntactic categories on the fully inflected lexical word derived by the lexical rule of the morphological processes in the lexicon.
Therefore, we will investigate how we adopt features for nouns such as cases including non-core and local cases such as \textsc{nom} (nominative),  \textsc{acc} (accusative), comparison (\textsc{cmpr}), and information structure  \textsc{top} (topic) (Table~\ref{morphological-schema-noun}). 
It will also include a typology of \texttt{jkb} (adverbial marker), which raises ambiguities. 
An adverbial marker can represent `dative' which marks the indirect object,
`instrumental' which marks means by which an action occurred, 
`allative' which marks a type of locative grammatical case, 
`ablative' which expresses motion away from something, or
`comparative' (\textsc{cmpr}, 예상\textit{yesang}.
We leave a detailed study on nouns and other grammatical categories for future work.
All datasets of K-UniMorph are available at \url{https://github.com/jungyeul/K-UniMorph} to reproduce the results. 
}


\section*{Acknowledgement}
We would like to thank Ekaterina Vylomova and Omer Goldman at the UniMorph project for their help and support.
We also wish to thank three anonymous reviewers for providing us with helpful feedback. 
This research was based upon work partially supported by the \textit{Students as Partners Course Design Grants} through the Office of the Provost \& Vice-President Academic at the University of British Columbia to Eunkyul Leah Jo, and by \textit{Basic Science Research Program} through the National Research Foundation of Korea (NRF) funded by the Ministry of Education (2021R1F1A1063474) to KyungTae Lim.
This research was also supported in part through computational resources and services provided by Advanced Research Computing at the University of British Columbia.


\appendix

\section{Neural Experiment Description}

We use the default setting of \texttt{fairseq} {for the neural experiment for the Table~\ref{experiment-results} in $\S$\ref{morphological-reinflection}} as described in Table~\ref{hyperparameter}.
\begin{description}
\item[\texttt{fairseq}] \texttt{fairseq-preprocess}, \texttt{fairseq-train} and \texttt{fairseq-interactive}.
\item[GPU] around 1 hour of GPU has been consumed for the training step for each experiment. 
\item[Total runtime] It takes about 2 to 3 hours for completing one experiment including all steps (preprocessing, training and evaluation). 
\item[Results] A single run with a seed number 

\end{description}

\begin{table}[ht]
    \centering
\begin{tabular}{l|c}\hline
task & translation \\
arch & transformer \\
dropout & 0.3 \\
learning rate & 0.0001  \\ 
lr-scheduler & inverse\_sqrt \\
attention-dropout & 0.3 \\
activation-dropout & 0.3 \\
activation-fn & relu \\
encoder-embed-dim & 256 \\
encoder-ffn-embed-dim & 1024 \\
encoder-layers & 4 \\
encoder-attention-heads & 4 \\
decoder-embed-dim & 256 \\
decoder-ffn-embed-dim & 1024 \\
decoder-layers & 4 \\
decoder-attention-heads & 4 \\
optimizer & adam \\
adam-betas & (0.9, 0.98) \\
clip-norm & 1.0 \\
warmup-updates & 4000 \\
label-smoothing & 0.1 \\
batch-size & 400 \\
max-update & 20000 \\\hline
\end{tabular}
    \caption{Hyperparameter}
    \label{hyperparameter}
\end{table}

\end{document}